\documentclass[10pt,twocolumn,letterpaper]{article}

\usepackage{
    wrapfig,
    lipsum,
    booktabs,
    placeins,
    float
}

\usepackage[pagenumbers]{iccv}
\definecolor{iccvblue}{rgb}{0.21,0.49,0.74}
\usepackage[pagebackref,colorlinks,allcolors=iccvblue]{hyperref}

\title{
RewardSDS: Aligning Score Distillation via Reward-Weighted Sampling
}

\author{
    \textbf{Itay Chachy} \quad \textbf{Guy Yariv} \quad \textbf{Sagie Benaim} \\
    Hebrew University of Jerusalem \\
    {\tt\small \{itay.chachy, guy.yariv, sagie.benaim\}@mail.huji.ac.il}
}

\begin{document}

\twocolumn[{
\renewcommand\twocolumn[1][]{#1}
\maketitle
\begin{center}
\vspace{-2.0em}
    \centering
    \captionsetup{type=figure}
\includegraphics[width=1.0\linewidth]{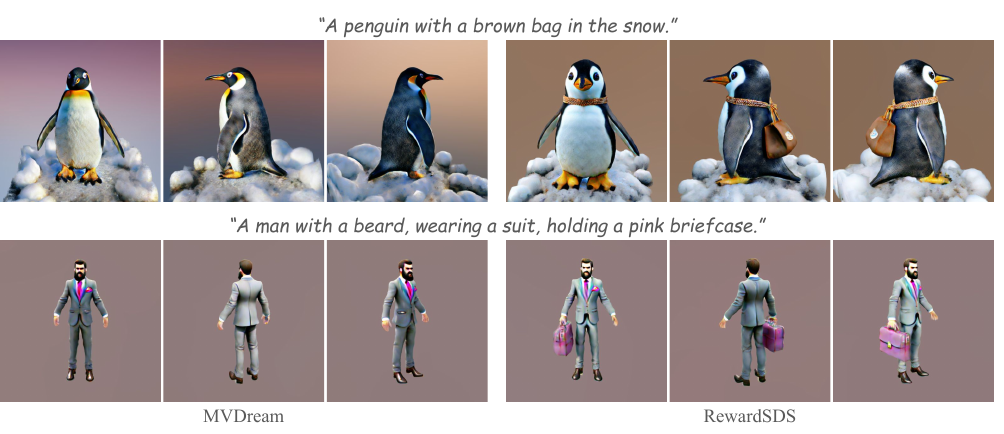}
    \vspace{-2.0em}
    \captionof{figure}{
     RewardSDS is a plug-and-play score distillation approach that allows for a reward-aligned generation. It can be applied to various tasks and extend diverse set of distillation approaches, boosting their performance and alignment. Here, we demonstrate it by replacing the standard SDS of the state-of-the-art MVDream~\cite{shi2024mvdreammultiviewdiffusion3d} approach with RewardSDS for text-to-3D generation. 
     }
\label{fig:teaser}
\end{center}
}]

\begin{abstract}
\label{sec:abstract}
\noindent\vspace{-1cm}

Score Distillation Sampling (SDS) has emerged as an effective technique for leveraging 2D diffusion priors for tasks such as text-to-3D generation. While powerful, SDS struggles with achieving fine-grained alignment to user intent. To overcome this, we introduce RewardSDS, a novel approach that weights noise samples based on alignment scores from a reward model, producing a weighted SDS loss. This loss prioritizes gradients from noise samples that yield aligned high-reward output. Our approach is broadly applicable and can extend SDS-based methods. In particular, we demonstrate its applicability to Variational Score Distillation (VSD) by introducing RewardVSD. We evaluate RewardSDS and RewardVSD on text-to-image, 2D editing, and text-to-3D generation tasks, showing significant improvements over SDS and VSD on a diverse set of metrics measuring generation quality and alignment to desired reward models, enabling state-of-the-art performance. Project page is available at \url{https://itaychachy.github.io/reward-sds/}.

\end{abstract}    
\section{Introduction}
\label{sec:intro}

Diffusion models have shown remarkable success in generating high-fidelity and diverse images ~\cite{rombach2022high, ramesh2022hierarchical, saharia2022photorealistic} and videos~\cite{blattmann2023stable, ho2022imagen, singer2022make}. However, their success often hinges on the availability of large-scale datasets, a requirement that poses a significant challenge in modalities like 3D content generation. In such data-scarce scenarios, leveraging diffusion models as priors becomes crucial, allowing one to improve generation quality. To this end, Score Distillation Sampling (SDS) was introduced~\cite{poole2022dreamfusiontextto3dusing2d, wang2023score}, distilling a pre-trained 2D diffusion model into a 3D representation by optimizing a loss that encourages rendered samples to have high scores under an underlying 2D diffusion model. While SDS has shown great promise, achieving fine-grained control and alignment with user intent remains a challenge. A different line of recent work has shown that one can utilize reward-based sample selection to align generative models such as LLMs~\cite{rafailov2023direct, brown2024large, snell2024scaling} or diffusion models~\cite{wallace2024diffusion, karthik2023if, liu2024correcting}. Motivated by these works, we ask: How can one effectively harness reward-based sample selection to align outputs produced by score distillation?

Building on SDS, several works have explored methods to improve the optimization process.  Variational Score Distillation (VSD)~\cite{wang2023prolificdreamerhighfidelitydiversetextto3d} optimized the 3D parameters as a random variable instead of a constant using a particle-based variational framework. 
Several other methods were proposed, either modifying or generalizing SDS~\cite{yu2023text, lee2024dreamflow, mcallister2025rethinking, zheng2025recdreamer, katzir2023noise}, or adapting it to the editing setting~\cite{hertz2023delta, koo2024posterior}. 
In the context of SDS alignment, recently, DreamReward~\cite{ye2024dreamreward} proposed to train a reward model on multi-view images, and subsequently use it to adjust the SDS score. In contrast, our method is compatible with any pre-trained reward model, including those trained on 2D images, eliminating the need for costly multi-view annotations. These works rely on sampling from an underlying pre-trained diffusion model and weighting the contribution of each sample equally. However, some noise samples may correspond to high-reward regions in the output space, while others may lead to low-reward regions. 
As such, we propose a novel adaptation of SDS, called RewardSDS.  We begin by rendering an image $x$ from an underlying model $\theta$.
We then draw $N$ noises corresponding to timestep $t$ from a Gaussian distribution, resulting in noisy samples $x^1_t, \dots x^N_t$.  Each noisy sample $x^i_t$ is assigned an \textit{alignment score}. This score is obtained by first denoising $x^i_t$ using the diffusion model, and then feeding the denoised sample into a given reward model to obtain a reward value, which we term the ``alignment score''. The overall loss is then given by a \textit{weighted sum} of the SDS losses computed for individual noisy samples $x^1_t, \dots x^N_t$, where the weight is derived from the alignment score. As our adaptation is general and can be broadly applied, we also demonstrate its applicability to Variational Score Distillation (VSD), referred to as RewardVSD.
We validate our approach using both RewardSDS and RewardVSD.
We first present a controlled study on zero-shot text-to-image generation using a diverse set of prompts from Drawbench~\cite{saharia2022photorealistic} and MS-COCO~\cite{lin2014microsoft}. We evaluate the generated outputs both in terms of image quality and text alignment using a diverse set of metrics.
Our evaluation is conducted across multiple pretrained reward models, demonstrating significantly improved performance and better alignment to each reward model. We demonstrate a similar improvement on 2D editing in comparison to Delta Denoising Score~\cite{hertz2023delta}. 

We then consider text-guided 3D generation. Specifically, we follow the state-of-the-art work of MVDream~\cite{shi2024mvdreammultiviewdiffusion3d} utilizing a diffusion model pretrained on muliview images for our prior. 
We subsequently optimize either a NeRF~\cite{mildenhall2020nerfrepresentingscenesneural} or a 3DGs~\cite{kerbl20233dgaussiansplattingrealtime} backbone using score distillation on this pretrained diffusion model, following the pipeline of DreamFusion~\cite{poole2022dreamfusiontextto3dusing2d} or DreamGaussian~\cite{tang2023dreamgaussian}, replacing the use of the SDS loss with RewardSDS. This results
in significantly improved performance. In particular, in comparison to MVDream, which uses a NeRF backbone with standard SDS, the use of RewardSDS instead significantly boosts performance and reward-model alignment. We also demonstrate the superiority of RewardVSD to standard VSD. Lastly, we provide an extensive ablation study and analyze the time vs performance tradeoff of our approach. 

\section{Related Work}
\label{sec:related_work}

\noindent \textbf{Diffusion Models Alignment} \quad
Fine-tuning diffusion models for alignment with human preferences has been extensively explored through methods such as reinforcement learning, which optimizes reward signals~\cite{black2023training}, direct gradient backpropagation from reward functions~\cite{xu2023imagerewardlearningevaluatinghuman}, direct preference optimization based on diffusion likelihood~\cite{wallace2024diffusion}, and stochastic optimal control frameworks~\cite{domingo2024adjoint}. 

Sample selection and optimization have also emerged as an important paradigm for improving diffusion model sample quality and alignment at test time.  Some methods focus on metric-guided selection, using random search guided by a reward model~\cite{karthik2023if, liu2024correcting}, or similarity to references~\cite{tang2024realfill, samuel2024generating}. 
 Recent work~\cite{ahn2024noise, zhou2024golden} approximate ``good" noise distributions with neural networks for efficient sampling. Unlike these works we consider alignment in the context of score distillation for tasks such as text-to-3D generation. 

\noindent \textbf{Score Distillation Sampling} \quad
Score Distillation Sampling (SDS)~\cite{poole2022dreamfusiontextto3dusing2d, wang2023score} has emerged as a essential technique for extending pre-trained 2D diffusion models to modalities like 3D content generation, where large-scale datasets are scarce. SDS leverages the 2D prior from text-to-image diffusion models to optimize rendering parameters, such as Neural Radiance Fields. However, the original SDS formulation exhibits limitations, notably artifacts like over-saturation and over-smoothing.  

Several works have aimed to improve the core distillation process with different strategies. Variational Score Distillation (VSD)~\cite{wang2023prolificdreamerhighfidelitydiversetextto3d} reformulates SDS as particle-based variational inference, optimizing a distribution of 3D scenes. Noise-Free Score Distillation~\cite{katzir2023noise} addresses over-smoothing by eliminating noise terms in the SDS objective. Classifier Score Distillation~\cite{yu2023text} re-evaluates the role of classifier-free guidance in score distillation. DreamFlow~\cite{lee2024dreamflow} suggests using a predetermined timestep schedule for distillation. RecDreamer~\cite{zheng2025recdreamer} proposed a 3D noise consistent distillation approach. JointDreamer~\cite{jiang2024jointdreamer} further improves multi-view consistency by using view-aware models within a joint distillation framework. SteinDreamer~\cite{wang2023steindreamer} employs Stein's identity to reduce gradient variance in SDS.
DreamTime~\cite{huang2023dreamtime} proposed an improved optimization strategy for score distillation.
Furthermore, works like Debiased-SDS~\cite{hong2023debiasing}, Perp-Neg~\cite{armandpour2023re}, DreamControl~\cite{huang2024dreamcontrol}, and ESD~\cite{wang2024taming} explore improved control over pose and biases in SDS-based generation.
SDS-Bridge~\cite{mcallister2025rethinking} presents a unified score distillation framework formulated as a bridge between distributions. 
Delta Denoising Score (DDS)~\cite{hertz2023delta} and Posterior Denoising Score~\cite{koo2024posterior} adapt SDS for image editing by modifying the gradient direction to better preserve input image details during editing. 
\textit{Beyond these core algorithmic improvements}, research has also explored diverse applications of SDS such as generating SVG graphics~\cite{iluz2023word, jain2023vectorfusion}, sketches~\cite{xing2023diffsketcher}, textures~\cite{li2024learning, metzer2023latent}, typography~\cite{iluz2023word}, and dynamic 4D scenes~\cite{bahmani20244d, singer2023text}.  Importantly, these advancements are orthogonal to our reward-based approach and offer potential for synergistic combination. 

In the context of SDS alignment, DreamReward~\cite{ye2024dreamreward} has recently proposed to align SDS by training a reward model to score a set of multiview images. They then use this model to adjust the SDS score provided by the underlying 2D diffusion model. Our method differs in two main aspects: (1). Our method can be aligned with any pretrained reward model, specifically a reward model trained to align 2D images as well as non-differentiable reward models. In contrast, DreamReward require the expensive annotation of scores for multiview data. (2). DreamReward still perform an expectation over all noises corresponding to a given timestep. Our method, on the other hand, weighs different noises according to their alignment score. As such, one can view our contribution as orthogonal. 

\begin{figure*}[t!]
    \centering
    \includegraphics[width=\linewidth]{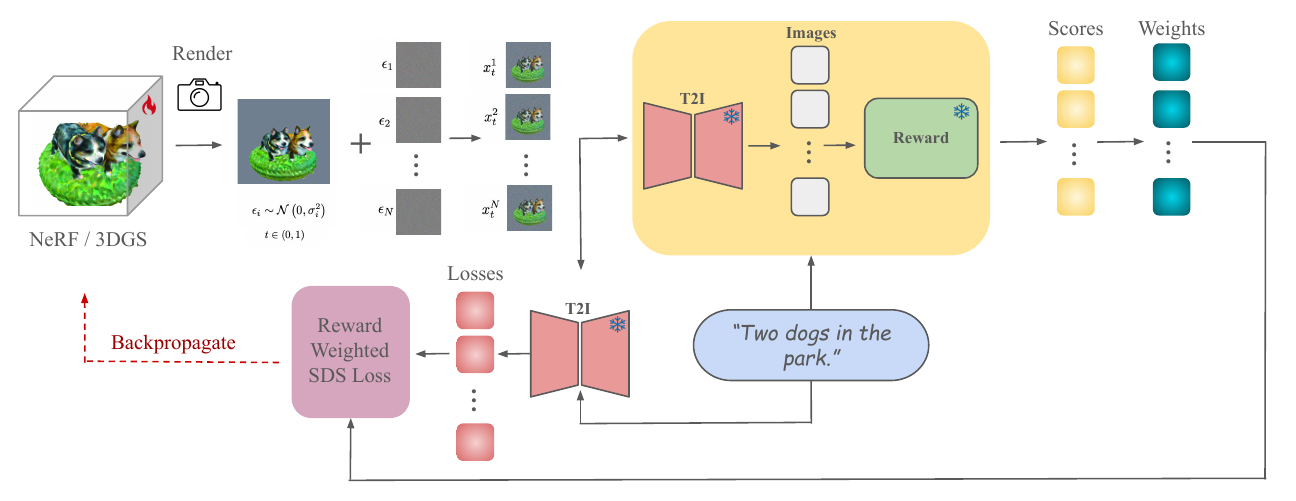}
    \caption{\textbf{RewardSDS illustration.} An image is first rendered from a given view and $N$ random noises are applied (at a given timestep). The noisy images are then scored by denoising them and applying a reward model on the output. These scores are then mapped to corresponding weights, which are used to weigh the contribution of each noisy sample in score distillation. }
    \label{fig:illustration}
    \vspace{-0.3cm}
\end{figure*}

\section{Method}
\label{sec:method}

We begin by outlining related background, specifically score distillation sampling (SDS) and variational score distillation (VSD). We then outline our approach as it applies to both SDS and VSD. An illustration is provided in Fig.~\ref{fig:illustration}. 

\noindent \textbf{Score Distillation Sampling (SDS)} \quad
The optimization process in SDS is derived from the training objective of diffusion models.  Given a clean image $x_0$ and a text prompt $y$, diffusion models are trained to predict the noise $\epsilon_t$ added at timestep $t$ to a noisy image $x_t$.  This training objective can be expressed as:
$$
L(x_0) = \mathbb{E}_{t\sim U(0,1), \epsilon_t \sim N(0, I)} \left[ w(t) \| \epsilon_\phi(x_t, y, t) - \epsilon_t \|_2^2 \right]
$$
where $\epsilon_\phi(x_t, y, t)$ is the noise predictor (typically a U-Net), $w(t)$ is a weighting function, and $x_t$ is obtained through the forward diffusion process:
\begin{align}
x_t = \alpha_t x_0 + \sigma_t \epsilon_t, \quad \epsilon_t \sim N(0, I) \label{eq:noisy_sample}
\end{align}
where $\alpha_t, \sigma_t$ are hyperparameters, chosen such that $\sigma_t^2 + \alpha_t^2 = 1$, and $\sigma_t$ gradually increases from 0 to 1.

In SDS, when $x_0 = g(\theta)$ is a rendered image from a differentiable generator $g$ parameterized by $\theta$, the parameters $\theta$ are updated by backpropagating the gradient of the loss:
$$
\nabla_\theta L_{SDS}(x_0 = g(\theta)) = \mathbb{E}_{t,\epsilon_t} \left[ w(t) (\epsilon_\phi(x_t, y, t) - \epsilon_t) \frac{\partial x_0}{\partial \theta} \right].
$$
This gradient update guides the parameters $\theta$ such that the rendered images $g(\theta)$ increasingly resemble samples from the text-conditioned 2D diffusion model.  Intuitively, SDS perturbs a rendered image $x_0$ by adding Gaussian noise to obtain $x_t$, then uses the pre-trained diffusion model to predict the noise $\epsilon_\phi(x_t, y, t)$ that should be present in $x_t$ to move it towards the real image distribution. The difference between the predicted noise and the added noise, weighted by $w(t)$ and backpropagated through $\frac{\partial x_0}{\partial \theta}$, provides an approximation of the gradient for optimizing $\theta$. 

\noindent \textbf{Variational Score Distillation (VSD)} \quad
Despite its effectiveness, SDS often suffers from issues like over-saturation and over-smoothing. To address these limitations and improve generation quality and diversity, Variational Score Distillation (VSD)~\cite{wang2023prolificdreamerhighfidelitydiversetextto3d} was proposed. VSD reformulates SDS within a particle-based variational inference framework. Instead of optimizing a single 3D scene representation, VSD optimizes a distribution $\mu(\theta|y)$ over scene parameters by introducing multiple particles $\{\theta_i\}_{i=1}^n$.  The VSD objective is defined as:
\begin{align}
\mu^* = \mathop{\arg\min}_\mu \mathbb{E}_{t} \left[ \frac{\sigma_t}{\alpha_t} \omega(t) D_{KL}(q^\mu_t(x_t|y) \| p_t(x_t|y)) \right]
\nonumber
\end{align}
where $q^\mu_t(x_t|y)$ is the distribution of rendered images from the particle set, and $p_t(x_t|y)$ is the target distribution from the pre-trained diffusion model.  To solve this optimization, VSD fine-tunes an additional U-Net, $\epsilon_\phi$, using LoRA, to estimate the score of the proxy distribution $q^\mu_t(x_t|y)$. That is $\epsilon_\phi$ is finetuned on rendered images  by $\{\theta^{(i)}\}_{i=1}^K$ with the standard diffusion objective:
\begin{align}
 \mathbb{E}_{t,\epsilon_t} \left[ w(t) ||\epsilon_\phi(x_t, y, t) - \epsilon_t)||^2_2\right] 
\end{align}
The gradient for each particle $\theta_i$ in VSD, $\nabla_{\theta_i} L_{VSD}(\theta_i)$ is then computed as:
\begin{align}
 \mathbb{E}_{t,\epsilon_t} \left[ w(t) (\epsilon_{pretrain}(x_t, y, t) - \epsilon_\phi(x_t, y, t)) \frac{\partial g(\theta_i)}{\partial \theta_i} \right]
\end{align}
where $\epsilon_{pretrain}$ is the original pre-trained diffusion denoiser, and $\epsilon_\phi$ is the fine-tuned score estimator. 

\noindent \textbf{RewardSDS}  \quad
Our core idea is to incorporate a reward model $R$ to guide the SDS optimization by prioritizing noise samples that are more likely to produce high-quality, aligned outputs. To this end, we rank a set of $N$ noise samples $\{\epsilon_{t}^{(i)}\}_{i=1}^N$ drawn at each iteration based on the reward scores they induce in the rendered image.  Specifically, for a given timestep $t$ and a set of noise samples $\{\epsilon_{t}^{(i)}\}_{i=1}^N \sim N(0, I)$, we generate a set of noisy images $\{x_{t}^{(i)}\}_{i=1}^N$ using Eq.~\ref{eq:noisy_sample}:
$$
x_{t}^{(i)} = \alpha_t x_0 + \sigma_t \epsilon_{t}^{(i)}, \quad i = 1, 2, ..., N,
$$
where $x_0 = g(\theta)$ is the rendered image. We then evaluate each noisy image $x_{t}^{(i)}$ by first denoising it and then using a reward model $R$ to obtain a set of reward scores $\{r^{(i)}\}_{i=1}^N$, where $r^{(i)} = R(x_{t}^{(i)})$. $R$ may optionally accept a text input. 

Based on these reward scores, the SDS loss for each noise sample $\epsilon_{t}^{(i)}$ is then weighted by a factor $w^{(i)}$ that is derived from $r^{(i)}$ (see Sec.~\ref{sec:experiments} for details). 

The Reward SDS loss is then computed as a weighted sum of the individual SDS losses, where $\{w^{(i)}\}_{i=1}^N$ are the weights assigned based on the reward ranking, and $w(t)$ is the standard SDS weighting function.  The gradient for updating the parameters $\theta$, $\nabla_\theta L_{R-SDS}(x_0 = g(\theta))$, is then given by:
\begin{align}
\mathbb{E}_{t} \left[ \frac{1}{N}\sum_{i=1}^N w^{(i)} \left[ w(t) (\epsilon_\phi(x_{t}^{(i)}, y, t) - \epsilon_{t}^{(i)}) \frac{\partial x_0}{\partial \theta} \right] \right] \label{eq:rewardsds}
\end{align}

\noindent \textbf{RewardVSD} \quad 
In RewardVSD, for each particle $\theta_i$ in the particle set $\{\theta_i\}_{i=1}^K$, we sample a set of $N$ noise samples $\{\epsilon_{t}^{(i,j)}\}_{j=1}^N$.  For each noise sample $\epsilon_{t}^{(i,j)}$, we compute the noisy image $x_{t}^{(i,j)}$ and its corresponding reward score $r^{(i,j)} = R(x_{t}^{(i,j)})$.  We then rank the noise samples $\{\epsilon_{t}^{(i,j)}\}_{j=1}^N$ based on their reward scores $\{r^{(i,j)}\}_{j=1}^N$ and assign weights $\{w^{(i,j)}\}_{j=1}^N$ accordingly.
The gradient for updating particle $\theta_i$, $\nabla_{\theta_i} L_{R-VSD}(\theta_i)$, is:
\begin{align}
\mathbb{E}_{t} \left[\frac{1}{N} \sum_{j=1}^N w^{(i,j)} \left[ w(t) (\epsilon_{pretrain}(x_{t}^{(i,j)}, y, t) \right. \right.  \\ 
 \left. \left. - \epsilon_\phi(x_{t}^{(i,j)}, y, t)) \frac{\partial g(\theta_i)}{\partial \theta_i} \right]\right] \nonumber
\end{align}

\section{Experiments}
\label{sec:experiments} 
We begin by evaluating our method on zero-shot text-to-image generation, reporting the effect of different reward models. Here, we directly optimize an image and compare the use of RewardSDS and RewardVSD to standard SDS and VSD, respectively. 
Second, we evaluate our approach on text-to-3D generation. 
Next we demonstrate the applicability of our approach for image editing by incorporating it into DDS~\cite{hertz2023deltadenoisingscore}, yielding RewardDDS. Lastly, we conduct an extensive ablation study and analyze optimization time vs performance tradeoff of using our method. Detailed implementation information for all experiments is provided in Appendix \ref{sec:impl_details}.

\noindent \textbf{Reward models.} \quad
Following \citet{ma2025inferencetimescalingdiffusionmodels}, we incorporate CLIPScore~\citep{radford2021learningtransferablevisualmodels} (using the ViT-L/14 variant), ImageReward~\citep{xu2023imagerewardlearningevaluatinghuman}, and the Aesthetic Score Predictor~\citep{schuhmann2022laion5bopenlargescaledataset} as optional reward models. We consider three separate model types, each optimized with respect to the reward produced by a different reward model. 
Each model serves a distinct role: CLIPScore validates text-image alignment by comparing visual and textual features. The Aesthetic Score Predictor assesses aesthetic quality, as it is trained to predict human ratings of synthesized images' visual appeal. ImageReward evaluates both alignment and aesthetics, learning general human preferences through a carefully curated annotation pipeline that includes ratings and rankings for text-image alignment, aesthetic quality, and harmlessness.

\noindent \textbf{Metrics.} \quad
As metrics, we utilize the three presented reward models. While each model type is optimized with respect to a specific reward model, we evaluate it with respect to all reward models. 
Additionally, we utilize Gemini 2.0 Flash~\footnote{\url{https://ai.google.dev/gemini-api}} as an LLM Grader, asking it to score each prompt-image pair on a scale from 1 to 10 based on criteria such as accuracy to the prompt, creativity and originality, visual quality and realism, and consistency and cohesion. We refer to this metric as the LLM Grader and present the average score. The exact prompt used as input for the LLM is sourced from~\citet{ma2025inferencetimescalingdiffusionmodels}, specifically in Fig. 16. To further validate our approach, we construct user studies, where we assess the realism of generated outputs and their alignment to the input text.

\subsection{Zero-Shot Text-to-Image Generation}
We begin by performing a zero-shot text-to-image generation where we optimize a latent map of size $64 \times 64 \times 4$ (corresponding to an image) in the Stable Diffusion's latent space~\citep{mcallister2024rethinkingscoredistillationbridge, katzir2023noisefreescoredistillation, wang2023prolificdreamerhighfidelitydiversetextto3d}. As opposed to text-to-3D, where the choice of model significantly impacts the result, this experiment allows us to measure the impact of our reward-based score distillation more directly. 

\noindent \textbf{Effect of different reward models.} \quad
First, we evaluate the effect of different reward models on our approach.
We randomly selected 25 prompts from the Drawbench benchmark~\citep{saharia2022photorealistic}, a diverse, general-purpose dataset of text prompts spanning multiple categories. We optimized RewardSDS and RewardVSD, using each of the three reward models noted above. 
Additionally, we report results for the SDS and VSD baselines alone.
As shown in Tab.~\ref{tab:reward_model_effect}, models trained on each of the three reward models effectively improve the baselines.
For both RewardSDS and RewardVSD, using the ImageReward model achieves the best overall results, attaining the highest LLM Grader score and the strongest performance on its own reward model. This suggests that ImageReward provides a well-balanced optimization that enhances alignment, aesthetics, and overall human preference.
Not surprisingly, in all cases, each reward model excels when evaluated with its corresponding reward model. However, we find that optimizing with respect to one reward model also results in an improved performance with respect to the other reward models. 

\begin{figure}[t!]
    \centering
    \includegraphics[width=\linewidth]{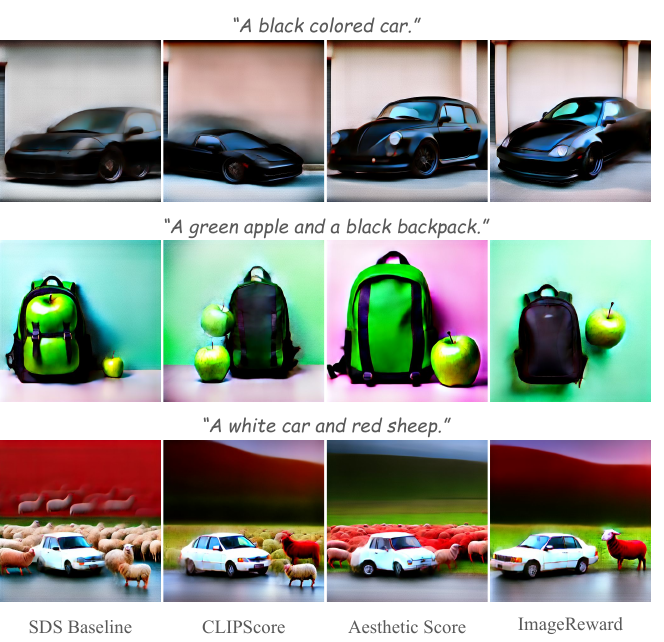}
    \caption{Qualitative comparison of generated outputs using different reward models for RewardSDS and the SDS baseline.}
    \label{fig:2d_reward_sds}
    \vspace{-0.3cm}
\end{figure}

Qualitative results showcasing RewardSDS with different reward models, along with the SDS baseline, are presented in Fig.\ref{fig:2d_reward_sds}, while results for RewardVSD with different reward models, along with the VSD baseline, are shown in Appendix \ref{sec:add_results_vsd_reward_models}. These examples illustrate the effect of integrating each reward model into image generation, along with the baseline.

\begin{table}[t!]
    \centering
    \small
    \setlength{\tabcolsep}{3.2pt} 
    \renewcommand{\arraystretch}{1.1} 
    \begin{tabular}{lcccc}
        \toprule
        Reward Model & CLIP$\uparrow$ & Aesthetic$\uparrow$ & ImageReward$\uparrow$ & LLM-G$\uparrow$ \\
        \midrule
        \multicolumn{5}{c}{RewardSDS} \\
        \midrule
        SDS Baseline & 27.93 & 5.42 & 0.59 & 6.74 \\
        \midrule
        CLIP        & \textbf{28.96} & 5.49 & 0.78  & 6.92 \\
        Aesthetic   & 27.30 & \textbf{5.67} & 0.64  & 6.83 \\
        ImageReward & 28.84 & 5.54 & \textbf{1.21} & \textbf{7.19} \\
        \bottomrule
        \multicolumn{5}{c}{RewardVSD} \\
        \midrule
        VSD Baseline & 27.41 & 5.15 & 0.51 & 6.73 \\
        \midrule
        CLIP & \textbf{27.99} & 5.21 & 0.73 & 6.87 \\
        Aesthetic & 27.37 & \textbf{5.35} & 0.55 & 6.74 \\
        ImageReward & 27.80 & 5.24 & \textbf{1.11} & \textbf{7.03} \\
        \bottomrule
    \end{tabular}
    \caption{
        Effect of different reward models on generated outputs using RewardSDS and RewardVSD. Each row represents results obtained by applying our method with a different reward model. The first row corresponds to the baseline SDS or VSD, where no reward model is used. 
        We report scores from three reward models: CLIP (CLIPScore), Aesthetic (Aesthetic Score), ImageReward, along with LLM-G (LLM Grader).
    }
    \label{tab:reward_model_effect}
    \vspace{-0.4cm}
\end{table}

\begin{figure}[t!]
    \centering
    \includegraphics[width=\linewidth]{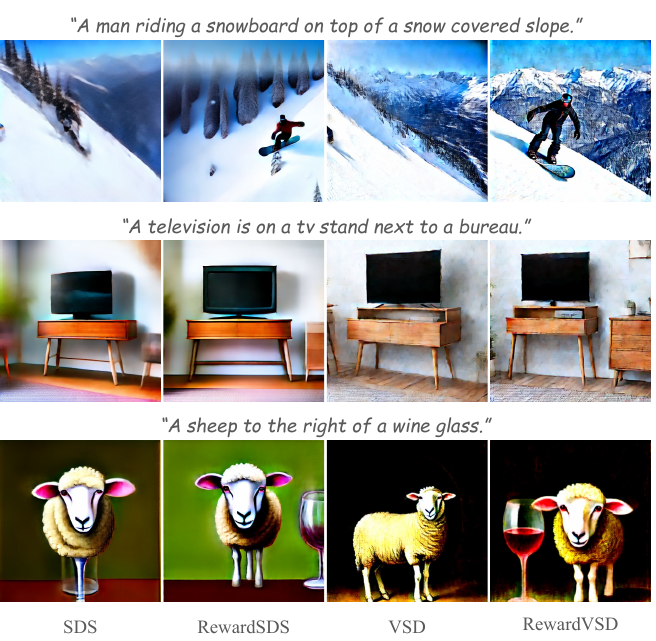}
    \caption{Qualitative comparison of zero-shot text-to-image generation using SDS, RewardSDS (ours), VSD,  RewardVDS (ours)}
    \label{fig:2d_generation}
    \vspace{-0.4cm}
\end{figure}

\noindent \textbf{Larger-scale comparison to baselines.} \quad
To further evaluate our approach, we conduct a larger-scale comparison to baselines. We randomly sample 100 prompts: 50 from MS-COCO~\citep{lin2015microsoftcococommonobjects} and 50 from Drawbench~\citep{saharia2022photorealistic}. We use ImageReward as the reward model and report results for CLIPScore, Aesthetic Score, and the LLM Grader. 
To verify that better alignment to the given reward model is not compensated by image realism or by alignment to the input text, we conduct a user study in which 50 users are asked to score from 1 (lowest) to 5 (highest): (1). How realistic is the generated image?, (2). How aligned is the generated image to the input text? 
For each of the 100 prompts used above, we generated corresponding images using SDS, VSD, RewardSDS and RewardVSD. Users are then presented the image corresponding for each method at random, and asked to rank each image from 1-5 on questions (1)-(2) above. 
As shown in Tab.~\ref{tab:2d_generation}, our reward-based sampling consistently improves generation performance across all metrics. 
Qualitative comparisons between RewardSDS, SDS, RewardVSD, and VSD are shown in Fig.~\ref{fig:2d_generation}.

\begin{table}[t!]
    \centering
    \small
    \setlength{\tabcolsep}{4pt}
    \renewcommand{\arraystretch}{1.1} 
    \begin{tabular}{lccccc}
        \toprule
        Method & CLIP$\uparrow$ & Aesthetic$\uparrow$ & LLM-G$\uparrow$ & Align.$\uparrow$ & Real.$\uparrow$ \\
        \midrule
        \multicolumn{6}{c}{MS-COCO} \\
        \midrule
        SDS         & 27.32 & 5.59 & 6.90 & 2.85 & 2.66 \\
        RewardSDS  & \textbf{27.86} & \textbf{5.84} & \textbf{7.10} & \textbf{3.93} & \textbf{3.18} \\
        \cmidrule(lr){1-6}
        VSD         & 27.40 & 5.39 & 6.85 & 3.15 & 2.31 \\
        RewardVSD  & \textbf{27.79} & \textbf{5.46} & \textbf{7.03} & \textbf{4.08} & \textbf{2.80}  \\
        \midrule
        \multicolumn{6}{c}{Drawbench} \\
        \midrule
        SDS         & 27.65 & 5.70 & 6.73 & 3.18 & 2.34 \\
        RewardSDS  & \textbf{28.29} & \textbf{5.86} & \textbf{7.11} & \textbf{3.48} & \textbf{2.66} \\
        \cmidrule(lr){1-6}
        VSD         & 26.74 & 5.32 & 6.64 & 2.85 & 2.34 \\
        RewardVSD  & \textbf{27.95} & \textbf{5.55} & \textbf{7.02} & \textbf{3.82} & \textbf{2.95} \\
        \bottomrule
    \end{tabular}
    \caption{
        Comparison of zero-shot text-to-image generation using RewardSDS/RewardVSD using ImageReward compared to SDS/VSD. We evaluate CLIPScore (CLIP), Aesthetic Score (Aesthetic), and LLM Grader (LLM-G). We also assess image alignment and realsim (user study MOS on a scale of 1 to 5). 
    }
    \label{tab:2d_generation}
    \vspace{-0.5cm}
\end{table}

\subsection{Text-to-3D Generation} 
\label{sec:3d_gen}
We now evaluate our approach on text-to-3D generation. 
Specifically, we follow the state-of-the-art work of MVDream~\cite{shi2024mvdreammultiviewdiffusion3d} by first training a multiview diffusion model. We subsequently optimize either a NeRF~\cite{mildenhall2020nerfrepresentingscenesneural} or a 3DGs~\cite{kerbl20233dgaussiansplattingrealtime} backbone using score distillation on this pretrained diffusion model, following the pipeline of DreamFusion~\cite{poole2022dreamfusiontextto3dusing2d} or DreamGaussian~\cite{tang2023dreamgaussian}, replacing the use of the SDS loss with RewardSDS.
3DGs was trained on 30 random prompts from the DreamFusion Gallery~\footnote{\url{https://dreamfusion3d.github.io/gallery.html}}, while NeRF was trained using 22 hand-crafted prompts. We refer readers to Appendix \ref{sec:nerf-prompts} for the full prompt list.
We measure CLIPScore, Aesthetic Score, and LLM Grader as automatic metrics of 10 randomly sampled views from each generated scene and report the average score.
We also consider a user study to assess alignment and realism, which follows the procedure of the 2D setting, but where we use 10 random views from each scene. 

\begin{figure}[t!]
    \centering
    \vspace{-0.3cm}
    \includegraphics[trim={0 0cm  0 0},clip, width=0.98\linewidth]{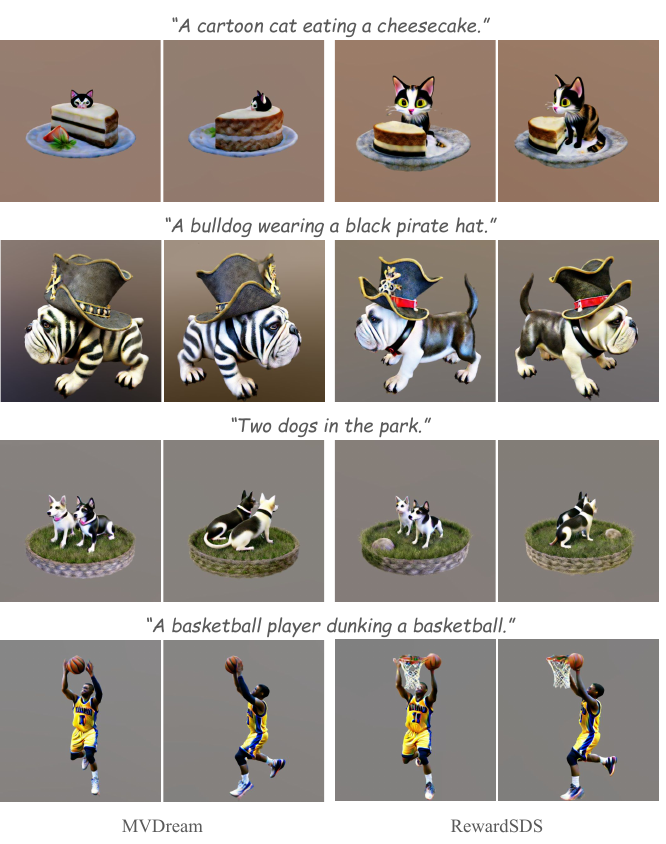} \\
    (NeRF Backbone)
    \\
    ~
    \includegraphics[trim={0 0cm  0 0},clip,width=2.20\linewidth]{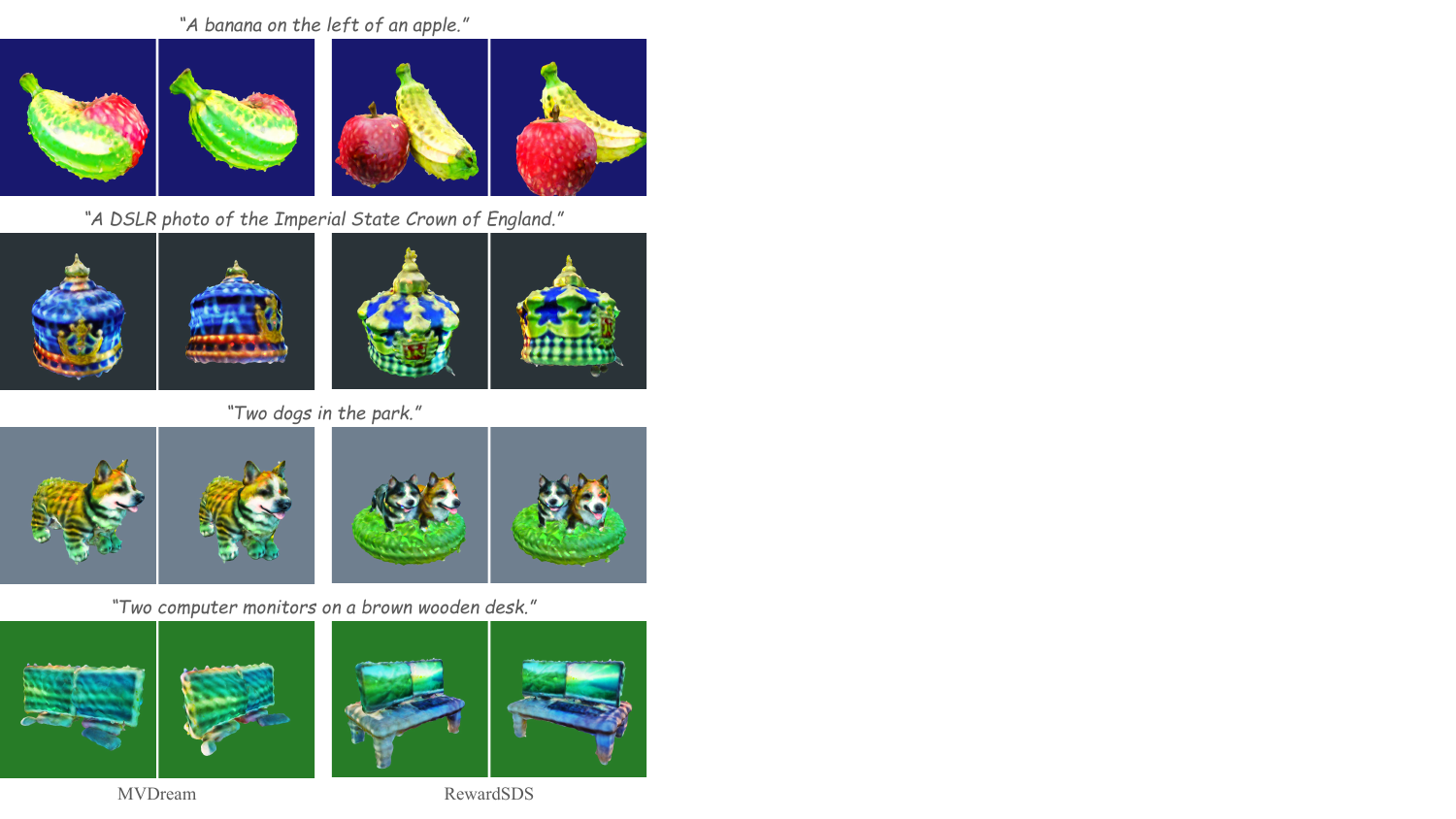} \\
    (3DGs Backbone)
    \\
    \caption{
        Qualitative comparison of text-to-3D generation based on (a) NeRF and (b) 3DGs, in comparison to MVDream.
    }
    \label{fig:3d_generation}
    \vspace{-0.8cm}
\end{figure}

\begin{figure}[t!]
    \centering
    \vspace{-0.3cm}
    \includegraphics[width=0.98\linewidth]{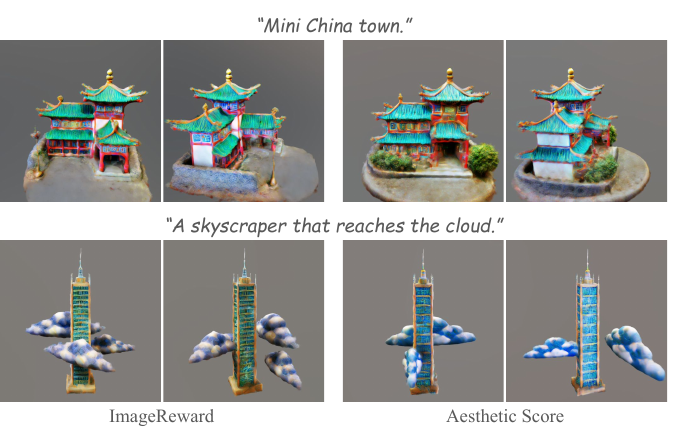}
    \caption{
        Qualitative comparison of text-to-3D generation using different reward models. We consider a NeRF backbone optimized with RewardSDS, either using the ImageReward reward model or Aesthetic Score reward model. As can be seen, using aesthetic reward results in adding bushes (top row) and a different (more aesthetic) color (both rows). 
    }
    \label{fig:3d_generation_rewards}
    \vspace{-0.3cm}
\end{figure}

As shown in Tab.~\ref{tab:3d_generation}, our method consistently improves all metrics using both 3D Gaussian Splatting and NeRF backbones, demonstrating its effectiveness in enhancing text-scene alignment and overall visual quality in 3D generation. Qualitative results, shown in Fig.\ref{fig:3d_generation}, illustrate the superiority of our method over the baseline in both settings. Additional examples can be found in the Appendix \ref{sec:add_results_3d}, specifically on the attached webpage. 
In addition, we qualitatively assess the effect of using different reward models. Specifically, in Fig.~\ref{fig:3d_generation_rewards} we demonstrate the effect of using RewardSDS with ImageReward in contrast to Aesthetic reward. 
In Appendix \ref{sec:add_results_3d}, we also provide additional results for standard DreamGaussian~\cite{tang2023dreamgaussian} training (without MVDream pertaining), using both RewardSDS and RewardVSD, showcasing our advantage to standard SDS and VSD. 

\begin{table}[t!]
    \centering
    \small
    \setlength{\tabcolsep}{2pt}
    \renewcommand{\arraystretch}{1.1} 
    \begin{tabular}{lccccc}
        \toprule
        Method & CLIP$\uparrow$ & Aesthetic$\uparrow$ & LLM-G$\uparrow$ & Align.$\uparrow$ & Real.$\uparrow$ \\
        \midrule
        \multicolumn{6}{c}{Gaussian Splatting} \\
        \midrule
        MVDream         & 24.71 & 5.73  & 4.90  & 2.74 & 2.28 \\
        RewardSDS  & \textbf{25.24} & \textbf{5.79}  & \textbf{5.31}  & \textbf{4.11} & \textbf{3.13} \\
        \midrule
        \multicolumn{6}{c}{NeRF} \\
        \midrule
        MVDream         &   26.19    &   5.83    & 5.86      & 3.51 & 3.14 \\
        RewardSDS  &  \textbf{27.12}     &    \textbf{5.97}   &  \textbf{6.07}     & \textbf{4.21} & \textbf{3.79} \\
        \bottomrule
    \end{tabular}
    \caption{
        Comparison of text-guided 3D generation using MVDream as our backbone, with and without reward-based sampling. We evaluate CLIPScore (CLIP), Aesthetic Score (Aesthetic), and LLM Grader (LLM-G). Additionally, we assess alignment and realism using a user study (MOS on a scale of 1 to 5). 
    }
    \label{tab:3d_generation}
    \vspace{-0.4cm}
\end{table}

\subsection{Image Editing}

We extend the editing method of DDS~\citep{hertz2023deltadenoisingscore}.

\begin{figure}[t!]
    \centering
    \includegraphics[width=\linewidth]{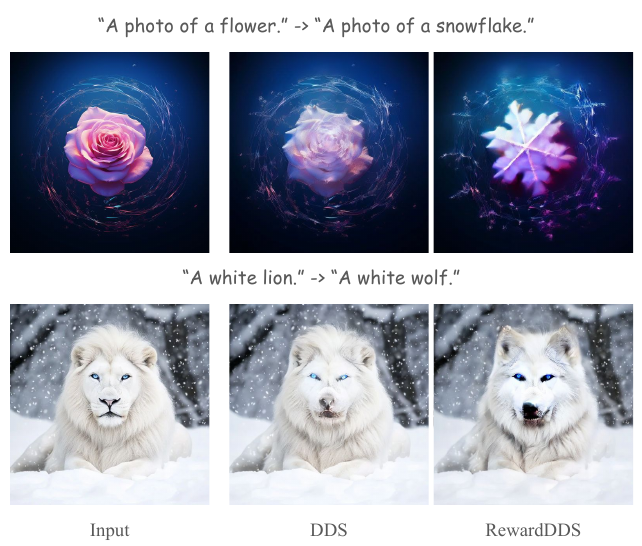}
    \caption{Qualitative comparison of image editing. We compare DDS to our adaptation, RewardDDS.}
    \label{fig:image_editing}
    \vspace{-0.4cm}
\end{figure}

DDS can be expressed as the difference of SDS of the source prompt and the SDS score of the target prompt. As such, our RewardDDS is simply the difference between two RewardSDS scores. We use five noise candidates and ImageReward as the reward model. 
We evaluate RewardDDS against DDS on 100 randomly sampled examples from the InstructPix2Pix dataset~\citep{brooks2023instructpix2pixlearningfollowimage}, pairing each target prompt with the generated image and computing automatic metrics (CLIPScore, Aesthetic Score, and LLM Grader). Tab.~\ref{tab:image_editing} shows that RewardDDS outperforms DDS across all metrics. A qualitative comparison is given in 
Fig.~\ref{fig:image_editing}.

\begin{table}[t!]
    \centering
    \small
    \setlength{\tabcolsep}{5pt}
    \begin{tabular}{lccc}
        \toprule
        Method & CLIPScore$\uparrow$ & Aesthetic Score$\uparrow$ & LLM Grader$\uparrow$ \\
        \midrule
        DDS & 24.00 & 5.77 & 6.89 \\
        RewardDDS & \textbf{24.19} & \textbf{5.80} & \textbf{7.09} \\
        \bottomrule
    \end{tabular}
    \caption{
    Quantitative comparison of image editing.}
    \label{tab:image_editing}
    \vspace{-0.4cm}
\end{table}

\subsection{Ablation Studies}

Our method allows for a diverse set of design choices, which we analyze here. We consider a common experimental setup for all ablations: We evaluate zero-shot text-to-image generation on 25 randomly sampled prompts from the Drawbench benchmark using ImageReward as the reward model and the LLM Grader as the primary metric. We use RewardSDS as our method, while the baseline is SDS.

\noindent \textbf{Noise weighting.} \quad
We ablate the 
choice of noise weighting ($w^{(i)}$),  applied to the $N$ candidate noise samples drawn at each iteration. 
Specifically, we compare the following strategies: (i) Random, where a weight of 1 is assigned to a randomly selected candidate and 0 to all others; (ii) Softmax, where a softmax function is applied to the reward scores;
(iii) Winner-takes-all, in which only the candidate with the highest reward is used (assigned a weight of 1, while the rest is assigned a weight of 0); 
\begin{wraptable}{r}{2.5cm}
\vspace{-0.0cm}
    \centering
    \small
    \setlength{\tabcolsep}{3pt}
    \renewcommand{\arraystretch}{1.1} 
    
    \begin{tabular}{lc}
        \toprule
        Scheme & LLM-G$\uparrow$ \\
        \midrule
        (i) & 6.69 \\
        (ii) & 7.08 \\
        (iii) & 7.05 \\
        (iv) & 7.13 \\
        (v) & 7.12 \\
        (vi) & \textbf{7.17} \\
        \bottomrule
    \end{tabular}
    \caption{Effect of different noise weighing schemes.}
    \label{tab:noise_weighing}
    \vspace{-0.3cm}
\end{wraptable}
(iv) Two winners-take-all (as in (iii) but with two highest-reward candidates); 
(v) Step towards best, away from worst, which takes a positive step toward the candidate with the highest reward (weight 0.9) while subtracting the influence of the lowest-reward candidate (weight -0.1); 
(vi) Step towards top-2, away from bottom-2, which is as in (v) but with two highest and two-lowest reward samples. 
Tab.~\ref{tab:noise_weighing} summarizes the performance of each scheme in terms of the LLM Grader score. 

\begin{figure}[t]
    \centering
    \includegraphics[width=\linewidth]{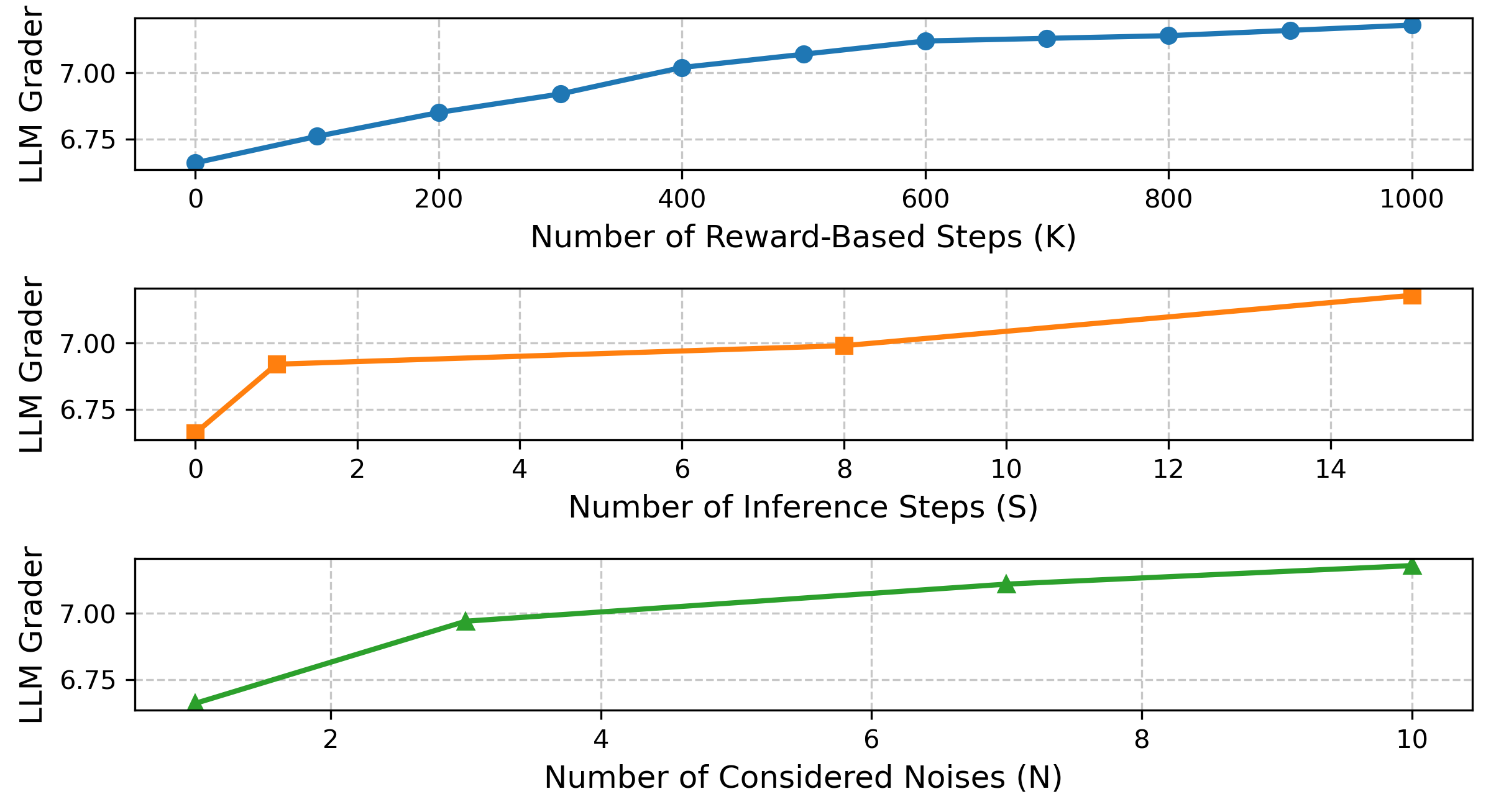}
    \caption{
        The top graph shows the impact of the number of reward-based steps ($K$), the middle graph presents the impact of the number of reward-based steps ($S$), and the bottom graph illustrates the impact of the number of considered noises ($N$).}
    \label{fig:ablation_scaling}
    \vspace{-0.5cm}
\end{figure}

\noindent \textbf{Reward-based optimization steps.} \quad
We now consider whether RewardSDS can be applied for a smaller number of steps $K$ out of the total number of optimization steps, while using standard SDS for the remaining steps. 
We vary $K$ from $0$ to $1000$ (out of a total of $1000$ optimization steps), in increments of $100$. As shown in Fig.~\ref{fig:ablation_scaling} (top graph), performance improves as $K$ increases, with noticeable saturation around $K \approx 600$ steps. 
Importantly, even for $K=100$, we observe superior results compared to $K=0$. 

\noindent \textbf{Effect of the number of noises sampled.} \quad
Next, we ablate the number of noise samples sampled (and weighted) at each optimization step, denoted as $N$ (as in Eq.~\ref{eq:rewardsds}). We test four different values of $N$: 1, 3, 7, and 10, where $N = 1$ serves as the baseline (no reward model). The results, presented in Fig.~\ref{fig:ablation_scaling} (middle graph), indicate that increasing $N$ consistently improves performance. Note the significant improvement already from $N = 1$ to $N = 3$. 
Qualitative results,
shown in Appendix \ref{sec:add_results_n_noises} further illustrate this. 

\noindent \textbf{Number of inference steps for noise selection.} \quad
We consider the effect of the number of denoising steps ($S$) used to denoise a noisy sample. 
We consider three different values: $S = 1$, $S = 8$, and $S = 15$, alongside a baseline without using the reward model ($S = 0$). As can be seen in Fig.~\ref{fig:ablation_scaling} (bottom graph), 
increasing $S$ consistently improves performance, with minimal increase in steps ($S = 1$) already outperforms the baseline.

\begin{figure}[t!]
    \centering
    \includegraphics[width=\linewidth]{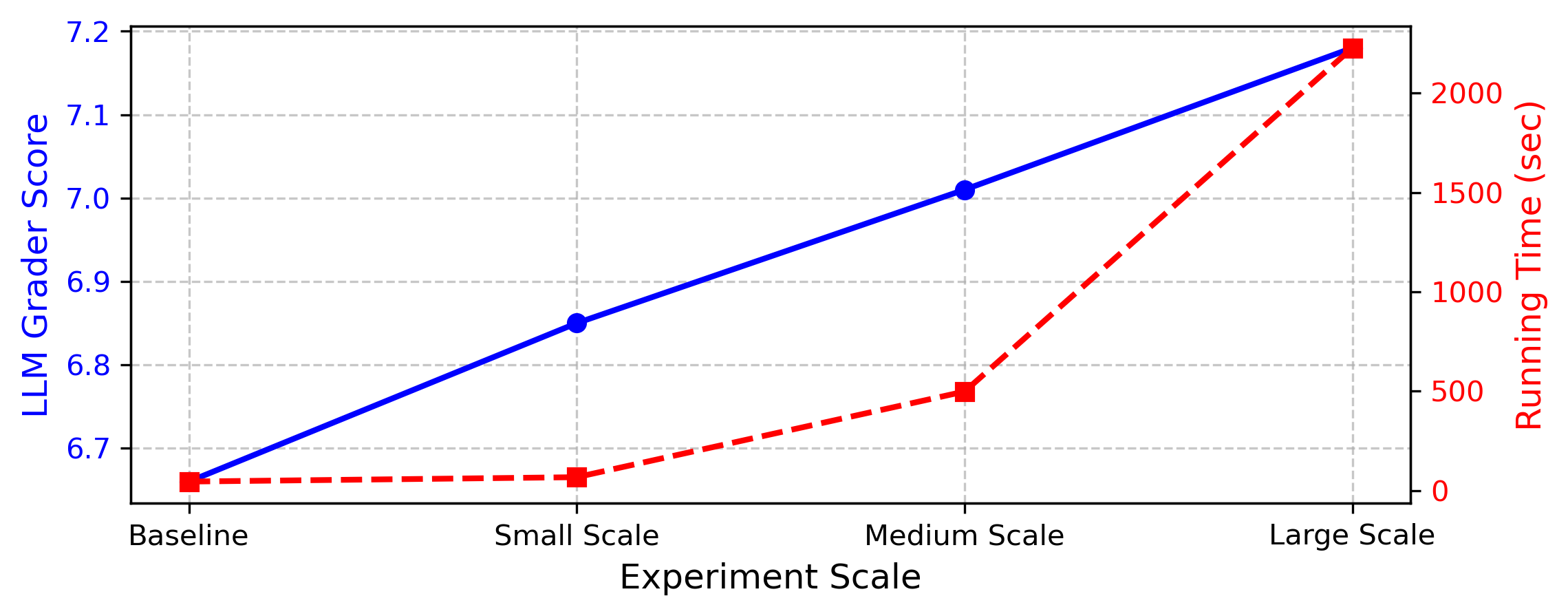}
    \caption{
        Tradeoff between running time and generation quality.  
    }
    \label{fig:time_vs_quality}
    \vspace{-0.2cm}
\end{figure}

\noindent \textbf{Time vs. quality tradeoff.} \quad
In the ablations above, by experimenting with $N$, $K$, and $S$ we observed two main conclusions: (i) increasing $N$, $K$, and $S$ improves performance, albeit at the cost of longer running time, and (ii) even small increases in these parameters already yield improvements over the baseline. Therefore, we further analyze the concrete tradeoff between quality and running time. Specifically, we evaluate four scenarios:
(i) Baseline – regular SDS, i.e., $N=1, K=0, S=0$. 
(ii) Small scale – minimal additional running time, i.e., $N=2, K=100, S=1$.  
(iii) Medium scale – with $N=5, K=500, S=8$.  
(iv) Large scale – with $N=10, K=1000, S=15$.  
In addition to reporting the LLM Grader's score, we also report the average running time required to optimize a single image (in seconds). The results, as shown in Fig.~\ref{fig:time_vs_quality}, suggest that the small-scale setting provides improvements over the baseline while maintaining a similar order of magnitude in running time (67 sec. vs. 45 sec.), while medium and large-scale settings significantly outperform the baseline in quality.

\noindent \textbf{Different SDS-based methods.} \quad 
Lastly, we demonstrate the versatility of our approach by showcasing its plug-and-play nature. Specifically, we integrate our method on SDS-Bridge~\cite{mcallister2024rethinkingscoredistillationbridge}, resulting in RewardSDS-Bridge. As SDS-Bridge adapts the input condition, our adaptation is otherwise the same as for SDS. 
We perform further evaluation via 
CLIPScore, Aesthetic Score, and the LLM Grader. Tab.~\ref{tab:sds_bridge} presents the results for SDS-Bridge and RewardSDS-Bridge, with our method consistently improving performance across all metrics. Qualitative comparisons can be found in Appendix \ref{sec:add_results_sds_bridge}.
\begin{table}[t!]
    \centering
    \small
    \setlength{\tabcolsep}{3pt}
    \renewcommand{\arraystretch}{1.1} 
    \begin{tabular}{lccc}
        \toprule
        Method & CLIPScore & Aesthetic Score & LLM Grader \\
        \midrule
        SDS-Bridge & 27.01 & 5.34 & 6.44 \\
        RewardSDS-Bridge & \textbf{27.94} & \textbf{5.58} & \textbf{7.31} \\
        \bottomrule
    \end{tabular}
    \caption{Comparison of SDS-Bridge and RewardSDS-Bridge on zero-shot text-to-image generation. 
    }
    \label{tab:sds_bridge}
    \vspace{-0.5cm}
\end{table}

\label{sec:ablations}

\section{Conclusion}
\label{sec:conclusion}

To conclude, we have presented RewardSDS, a novel approach for addressing the critical challenge of aligning score distillation with user intent through reward-weighted noise sample selection.  Across extensive evaluations encompassing zero-shot text-to-image generation, text-to-3D creation, and image editing, RewardSDS, and its VSD variant, RewardVSD, consistently and significantly outperformed standard SDS and VSD baselines across diverse metrics, including CLIPScore, Aesthetic Score, ImageReward, LLM Grader assessments, and user studies, with a model based on ImageReward notably demonstrating robust performance.  The general nature of RewardSDS facilitates seamless integration with various SDS extensions and pre-trained reward models, offering a flexible framework for improved alignment. 

{
    \small
    \bibliographystyle{ieeenat_fullname}
    \bibliography{main}
}

\clearpage
\appendix

\section{Additional Results}
\label{sec:add_results} 
\subsection{Qualitative Comparison of the Effect of Different Reward Models on RewardVSD}
\label{sec:add_results_vsd_reward_models} 
\begin{figure}[h!]
    \centering
    \includegraphics[width=\linewidth]{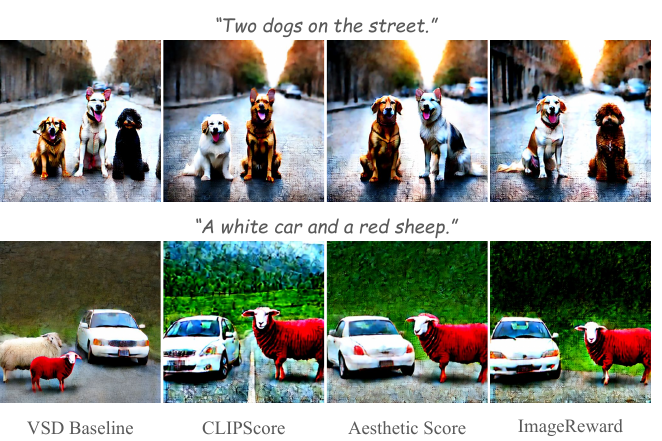}
    \caption{
        Qualitative comparison of generated outputs using different reward models for RewardVSD and the VSD baseline. Each row corresponds to a different reward model, with the input prompts shown at the bottom, taken from Drawbench.
    }
    \label{fig:2d_reward_vsd}
\end{figure}

To better understand the effect of different reward models in the RewardVSD framework, we provide a qualitative comparison. \cref{fig:2d_reward_vsd} illustrates the generated outputs using different reward models and the VSD baseline.

\begin{figure}[h!]
    \centering
    \includegraphics[width=\linewidth]{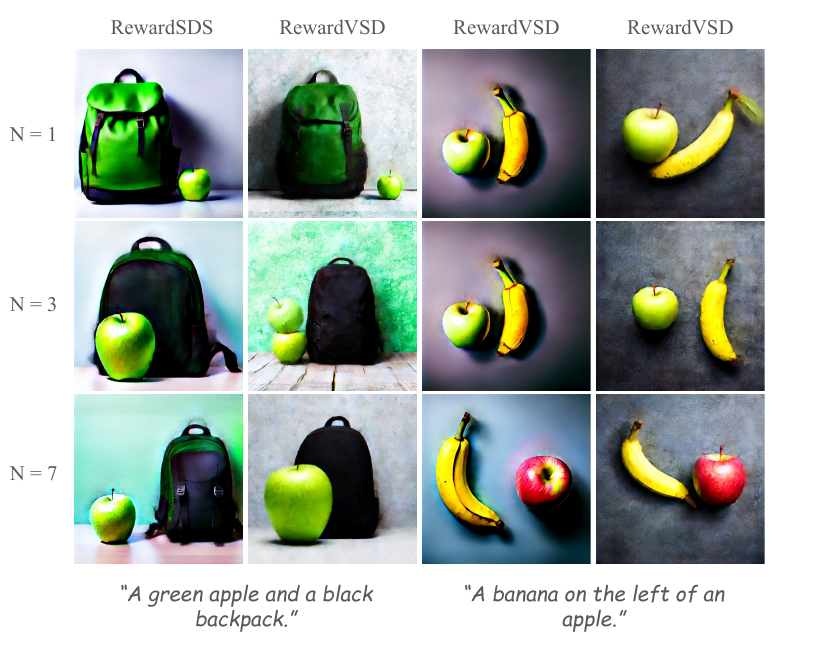}
    \caption{
        Qualitative results illustrating the effect of the number of considered noises ($N$). The top row presents the baseline method, while the input prompts are displayed below the images.}
    \label{fig:ablation_n_q}
\end{figure}

\begin{table}[t!]
    \centering
    \small
    \setlength{\tabcolsep}{2pt}
    \renewcommand{\arraystretch}{1.1}

    \begin{tabular}{lccc}
        \toprule
        Method & CLIPScore & Aesthetic & LLM Grader \\
        \midrule
        SDS         & 21.49 & 5.34 & 3.85 \\
        RewardSDS  & \textbf{22.87}& \textbf{5.53}  & \textbf{4.29} \\
        \midrule
        VSD         & 21.38 & 5.14 & 3.67 \\
        RewardVSD  & \textbf{22.03} & \textbf{5.41} & \textbf{4.11} \\
        \bottomrule
    \end{tabular}
        \caption{Comparison of text-guided 3D generation using standard SDS and VSD (with stable-diffusion-2.1-base), with and without reward-based sampling, over 30 randomly sampled prompts from DreamFusion gallery, with each score averaged over 10 randomly rendered views.}
    \label{tab:3d_generation_suppl}
\end{table}

\begin{figure*}[t!]
    \centering
    \includegraphics[width=\linewidth]{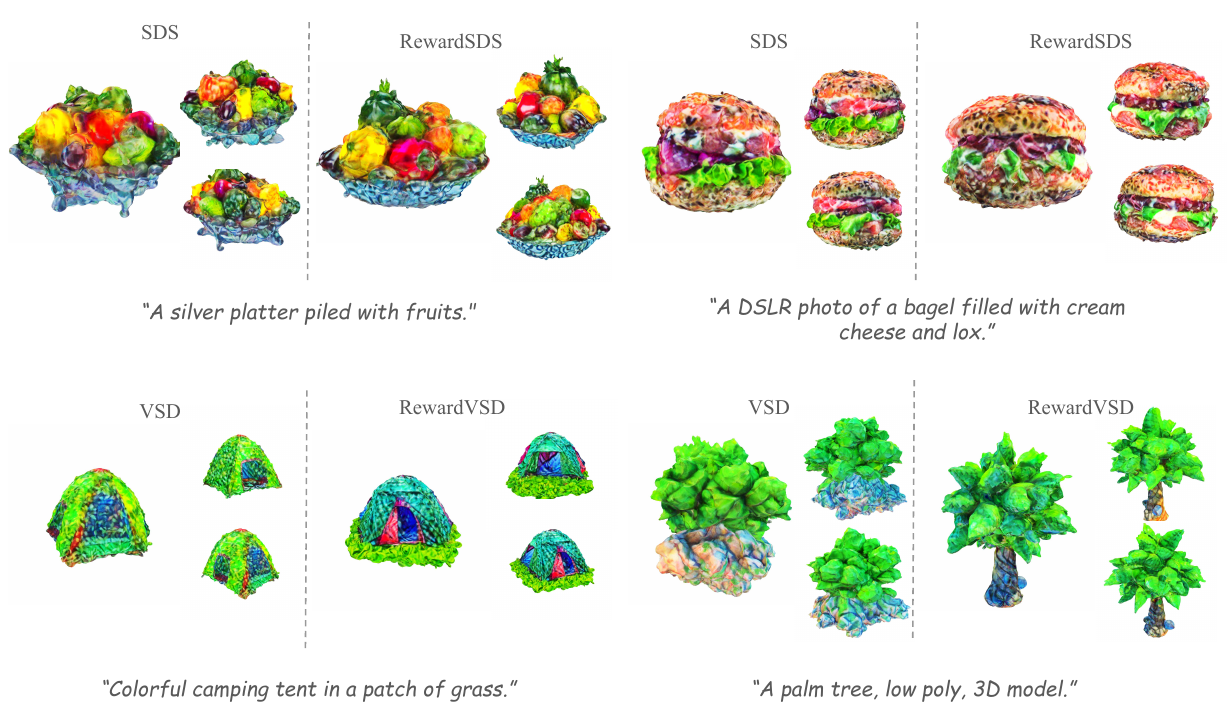}
    \caption{
        Qualitative results illustrating 
        text-to-3D results for SDS compared to RewardSDS and for VSD compared to our RewardVSD (without MVDream's pretraining).}
    \label{fig:additional_3d}
\end{figure*}

\subsection{Qualitative Comparison of the Effect of Number of Considered Noises}
\label{sec:add_results_n_noises} 
The number of noise samples ($N$) considered at each optimization step plays a crucial role in guiding the generation process. As discussed in the ablation studies, increasing $N$ leads to better alignment with the desired reward model and enhances the overall image quality. \cref{fig:ablation_n_q} provides qualitative examples illustrating how different values of $N$ affect the final generated outputs. As shown, larger values of $N$ result in more refined and coherent images, whereas smaller values introduce more variability and potential misalignment.

\subsection{Additional Text-to-3D results}
\label{sec:add_results_3d} 
Tab.~\ref{tab:3d_generation_suppl} provides a comparison of text-guided 3D generation using standard SDS and VSD (with stable-diffusion-2.1-base) in comparison to RewardSDS and RewardVSD, over 30 randomly sampled prompts from DreamFusion gallery, with each score averaged over 10 randomly rendered views. Unlike the main text, here we train on a standard diffusion model, not pretrained on multiview images as in MVDream. 
In Fig.~\ref{fig:additional_3d}, we provide corresponding text-to-3D results for SDS compared to RewardSDS and VSD compared to our RewardVSD.

\begin{figure}[t!]
    \centering
    \includegraphics[width=\linewidth]{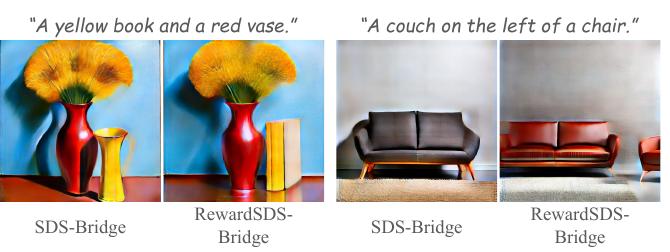}
    \caption{
    Qualitative comparison between SDS-Bridge and RewardSDS-Bridge. 
    }
    \label{fig:bridge}
\end{figure}

\subsection{Qualitative Comparison of RewardSDS-Bridge vs. SDS-Bridge}
\label{sec:add_results_sds_bridge} 
Building upon the comparisons presented in the ablation study, we provide further qualitative results comparing RewardSDS-Bridge with SDS-Bridge in Fig.~\ref{fig:bridge}. These results highlight the improved sample quality and alignment with text conditions achieved by our approach.

\section{Implementation details.}
\label{sec:impl_details} 
All experiments, including both our method and the baselines, we use a single L40s GPU.
For all zero-shot text-to-image, and image editing experiments, optimization is performed using the ADAM optimizer with a learning rate of 0.01. As the text-to-image model, we employ Stable Diffusion 2.1 Base~\citep{rombach2022highresolutionimagesynthesislatent}, setting the classifier-free guidance (CFG) scale to 100 for SDS-based experiments and 7.5 for VSD-based experiments. Each 2D image is generated in 1,000 optimization steps using $N=10$ (the number of noise samples drawn at each iteration), $K=1000$ (the number of reward-based optimization steps used during generation), and $S=15$ (the number of inference steps applied during noise selection to obtain refined reward estimates). The weighting strategy assigns a weight of 0.9 to the two highest scoring samples, a weight of -0.1 to the two lowest scoring samples, and 0 to the rest. With those settings, image optimization takes 2227 seconds. For image editing experiments, generation is performed in 200 steps, taking 211 seconds per image, and we use $N=5$, $K=200$, and $S=1$ as the hyperparameters, with the same weighting strategy as above.
For text-guided 3D generation, MVDream~\cite{shi2024mvdreammultiviewdiffusion3d} is used as our text-to-image prior, and we employ the public implementations of DreamGaussian~\cite{tang2023dreamgaussian} for 3DGs backbone optimization (leaving their second stage of texture refinement as is) and MVDream~\cite{threestudio2023} (without shading) for NeRF based optimization. 
We use the same settings as in the public implementations, and in NeRF training, we apply our method only in the first half of the optimization as we found this is sufficient for convergence with our method. Regarding the weighting strategy, we use the same one as in the 2D. Optimization of 3DGs takes 60 minutes and NeRF takes 10 hours.

\section{Hand-Crafted Prompts for NeRF-Based MVDream Evaluation}
\label{sec:nerf-prompts}

As described in Section~\ref{sec:3d_gen}, to evaluate NeRF-based 3D generation using MVDream, with and without our method, we utilized a set of 22 hand-crafted prompts. These prompts were carefully designed to cover a diverse range of objects, scenes, and subjects, ensuring a comprehensive assessment of text-to-3D generation quality. The full list of prompts is provided in Table~\ref{tab:nerf-training-prompts}. 

\begin{table}[H]
    \centering  
    \small
    \setlength{\tabcolsep}{6pt} 
    \renewcommand{\arraystretch}{1.2} 
    \begin{tabular}{r l}
        \toprule
        \# & Prompt \\ 
        \midrule
        1  & A basketball player dunking a basketball \\
        2  & A basketball player in a red jersey, high resolution, 4K \\
        3  & A bulldog wearing a black pirate hat \\
        4  & A cartoon cat eating a cheesecake, realistic \\
        5  & a DSLR photo of a ghost eating a hamburger \\
        6  & A guitar player on stage, high quality, realistic, HD, 8K \\
        7  & A man with a beard, wearing a suit, holding a pink briefcase, high quality, realistic, HD \\
        8  & A penguin with a brown bag in the snow \\
        9  & A man with a red scarf, highly detailed, 4K \\
        10 & A shitzu with a bowtie, high quality, realistic, HD, 8K \\
        11 & A skyscraper that reaches the clouds, high quality, realistic \\
        12 & A tiger in the jungle, high quality, realistic, HD \\
        13 & A white sofa next to a brown wooden table \\
        14 & A young girl flying a kite, high quality \\
        15 & An astronaut riding a horse \\
        16 & Argentinian football player, celebrating a goal, HD \\
        17 & Corgi riding a rocket \\
        18 & King Kong climbing the Empire State Building \\
        19 & Mini China town, highly detailed, 8K, HD \\
        20 & Red drum set, high quality, realistic, HD, 8K \\
        21 & Two dogs in the park \\
        22 & World cup trophy, high quality, realistic, HD \\
        \bottomrule
    \end{tabular}
    \caption{List of hand-crafted prompts used to evaluate NeRF-based MVDream with and without RewardSDS.}
    \label{tab:nerf-training-prompts}
\end{table}

\end{document}